\documentclass[letterpaper]{article} % DO NOT CHANGE THIS
\usepackage[]{aaai25}  % DO NOT CHANGE THIS
\usepackage{times}  % DO NOT CHANGE THIS
\usepackage{helvet}  % DO NOT CHANGE THIS
\usepackage{courier}  % DO NOT CHANGE THIS
\usepackage[hyphens]{url}  % DO NOT CHANGE THIS
\usepackage{graphicx} % DO NOT CHANGE THIS
\def\UrlFont{\rm}  % DO NOT CHANGE THIS
\usepackage{natbib}  % DO NOT CHANGE THIS AND DO NOT ADD ANY OPTIONS TO IT
\usepackage{caption} % DO NOT CHANGE THIS AND DO NOT ADD ANY OPTIONS TO IT
\usepackage{algorithm}
\usepackage{algorithmic}

\usepackage{newfloat}
\usepackage{listings}
\DeclareCaptionStyle{ruled}{labelfont=normalfont,labelsep=colon,strut=off} % DO NOT CHANGE THIS
\floatstyle{ruled}
\newfloat{listing}{tb}{lst}{}
\floatname{listing}{Listing}
\usepackage{xcolor}
\usepackage{amsmath}
\usepackage{tikz}
\usepackage{mathdots}
\usepackage{yhmath}
\usepackage{cancel}
\usepackage{color}
\usepackage{siunitx}
\usepackage{array}
\usepackage{multirow}
\usepackage{amssymb}
\usepackage{gensymb}
\usepackage{tabularx}
\usepackage{extarrows}
\usepackage{booktabs}
\usetikzlibrary{fadings}
\usetikzlibrary{patterns}
\usetikzlibrary{shadows.blur}
\usetikzlibrary{shapes}
\usepackage{xfrac}
\usepackage{subcaption}
\usepackage{hyperref}

\usepackage{amsmath, amssymb, amsthm}

\newcommand{\reals}[0]{\mathbb{R}}
\newcommand{\funcdef}[3]{#1\colon#2\to#3}

\DeclareMathOperator{\codec}{\textsc{Codec}}
\DeclareMathOperator{\kmac}{\textsc{KMAc}}

\title{Dimension Reduction for Symbolic Regression}
\author {
    Paul Kahlmeyer,
    Markus Fischer,
    Joachim Giesen
}
\affiliations {
    Friedrich Schiller University Jena\\
    paul.kahlmeyer@uni-jena.de, markus.fischer@uni-jena.de, joachim.giesen@uni-jena.de 
}

\begin{document}

\maketitle

% Abstract:
% 1. Symbolic Regression aims at recovery
% 2. Recovery of complex formulas is more challenging
% 3. Fixed combinations -> Substitutions -> Reduce complexity of expression
% 4. Here we adress... wie gehabt

\begin{abstract}
Solutions of symbolic regression problems are expressions that are composed of input variables and operators from a finite set of function symbols. One measure for evaluating symbolic regression algorithms is their ability to recover formulae, up to symbolic equivalence, from finite samples. Not unexpectedly, the recovery problem becomes harder when the formula gets more complex, that is, when the number of variables and operators gets larger. Variables in naturally occurring symbolic formulas often appear only in fixed combinations. This can be exploited in symbolic regression by substituting one new variable for the combination, effectively reducing the number of variables. However, finding valid substitutions is challenging. Here, we address this challenge by searching over the expression space of small substitutions and testing for validity. The validity test is reduced to a test of functional dependence. The resulting iterative dimension reduction procedure can be used with any symbolic regression approach. We show that it reliably identifies valid substitutions and significantly boosts the performance of different types of state-of-the-art symbolic regression algorithms\footnote{\href{https://github.com/kahlmeyer94/DAG\_search}{\textcolor{blue}{\texttt{https://github.com/kahlmeyer94/DAG\_search}}}}.
\end{abstract}

%------------------------------------------------------
% Introduction
%------------------------------------------------------
\section{Introduction}

In regression, we are given $n$ observations 
\[
\big(x^{(1)},y^{(1)}\big),\ldots,\big(x^{(n)},y^{(n)}\big)
\]
of $d$ input variables $x\in\mathbb{R}^d$ and one output variable $y\in\mathbb{R}$. The goal is to use the observations to find a function $f\colon \mathbb{R}^d\rightarrow \mathbb{R}$ that generalizes well beyond the given observations. Symbolic regression adds another goal, namely, to find an interpretable function $f$. In symbolic regression, $f$ is taken from a space of symbolic expressions that are composed of the input variables $x_1,\ldots,x_d$ and a finite set of function symbols, such as, for instance, 
\[
\{+,\, -,\, \cdot,\, /,\, \sqrt,\, \log,\, \exp,\, \sin,\, \cos,\, \texttt{{\small const\_expr}}\}.
\]
Washburn's equation
\[
L = \sqrt{\frac{\gamma rt \cos (\phi)}{2\eta}}
\]
is an illustrative example out of a set of 880 formulas that were extracted from Wikipedia articles by~\citet{bayesianscientist_guimera20}. Washburn's equation describes the penetration length ($L$) of a liquid into a capillary pore as a function of time ($t$), properties of the liquid, namely, its dynamic viscosity ($\eta$) and its surface tension ($\gamma$), and the geometry of the set-up, that is, the pore's radius ($r$) and the contact angle ($\phi$) between the pore and the liquid. A key insight from Washburn's equation is that the penetration length grows proportionally to the square root of the time variable, and the factor of proportionality is a function of the liquid's properties and the geometry of the set-up.  The right-hand side of Washburn's equation, short Washburn's formula, is a valid symbolic expression that is composed of five input variables 
\[
x_1 =t, x_2 =\eta, x_3=\gamma,x_4=r, \textrm{ and } x_5=\phi,
\]
and the function symbols $\cdot,\, /,\, \sqrt,\, \cos$, and $\texttt{{\small const\_expr}}$, where the multiplication symbol is used three times and $\texttt{{\small const\_expr}}$ is instantiated by the constant $2$. Washburn's formula is already quite complex. Its expression tree has five input variable nodes and eight operator nodes that are labeled by function symbols, that is, 13 nodes in total, a number that serves as a measure of complexity. 
\begin{figure}[h!]
    \centering
    \includegraphics[width = 0.8\columnwidth]{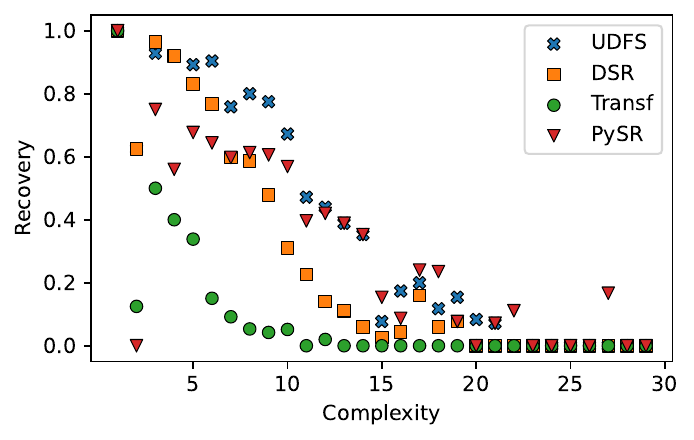}
    \caption{Fraction of successfully recovered expressions from a set of 880 formulas from Wikipedia's list of eponymous equations \citep{bayesianscientist_guimera20} for the state-of-the-art symbolic regressors \textsc{UDFS}: search over expression DAGs, \textsc{DSR}: reinforcement learning, \textsc{transf}: seq2seq learning, and \textsc{PySR}: genetic programming.}
    \label{fig:freq_intro}
\end{figure}

Figure~\ref{fig:freq_intro} shows, that the ability of state-of-the-art regressors to recover formulas from sampled data decreases when the complexity of the formulas grows. As suggested by \citet{srbench_lacava21}, the complexity of a formula is measured by the size of a corresponding minimal expression tree.
It turns out that none of the different, state-of-the-art symbolic regressor algorithms from Figure~\ref{fig:freq_intro} can recover Washburn's formula from observational data.  

We can simplify Washburn's formula, for instance, by representing the geometric set-up that is given by $r$ and $\phi$ in one variable 
$g(r,\phi) = r\cos(\phi)$. The size of the expression tree for the simplified Washburn formula, where the input variable nodes for $r$ and $\phi$ are substituted by an input variable node for $g(r,\phi)$, is reduced from twelve to nine, rendering the symbolic regression more tractable. After the substitution, the \textsc{UDFS} regressor from Figure~\ref{fig:freq_intro} can recover Washburn's formula from the observational data. Obviously, more aggressive simplifications are possible. However, since we are only given observations for the input variables and the corresponding output variables, valid substitutions are not obvious and finding them becomes a task of its own.

\citet{feynmanAI2_udrescu20} were first to suggest predicates for valid substitutions in their symbolic regression framework AIFeynman. The predicates, however, work only for a limited set of substitutions $g(x_i,x_j)$, namely differences $x_i-x_j$, sums $x_i+x_j$, products $x_i\cdot x_j$, and quotients $x_i/ x_j$ of input variables $x_i$ and $x_j$. The predicate for the substitution $g(x_i,x_j)=x_i-x_j$ works by comparing 
\[ 
f(\ldots, x_i,\ldots x_j,\ldots) \:\textrm{ with }\: f(\ldots, x_i+a,\ldots x_j+a,\ldots)
\]
for different values for $x_i, x_j\in\mathbb{R}$ and offsets $a\in\mathbb{R}$. If $g(x_i,x_j)$ is a valid substitution, then the difference of the two function values should be very small. The predicates for the remaining three substitutions work similarly.

Here, we generalize the ideas of \citet{feynmanAI2_udrescu20} to much more general substitutions. We use the framework of \citet{kahlmeyer2024udfs} to enumerate expression DAGs up to a certain complexity. Every DAG represents a possible substitution. For checking if a possible substitution is a valid substitution, we build on recent work by \citet{chatterjee:2020} and generalizations by \citet{azadkia:2021} and \citet{deb:2020} for certifying functional dependence from a sample. The fundamental assumption behind a regression problem is that the observations represent a functional dependence between the input variables $x_1,\ldots,x_d$ and the output variable $y$. Let $x_I$ be a subset of the input variables and $g(x_I)$ a possible substitution for $x_I$. If $g$ is valid, then the observations should still represent a functional dependence if we substitute $g(x_I)$ for $x_I$ in every data point.

Our substitution finding approach can be used together with any symbolic regression algorithm. In fact, as we show in the experimental section of this paper, it reduces the number of variables in the formulas of an established benchmark data set~\citep{srbench_lacava21} by $\sim$50\% and thereby significantly boosts the recovery performance of many conceptually fairly different, state-of-the-art symbolic regression algorithms    

\paragraph{Outline.}

In Section~\ref{sec:substitutions} we formally define substitutions, and review the recent progress in functional dependence measures in Section~\ref{sec:funcDependence}. Then, in Section~\ref{sec:loop}, we use the framework of \citet{kahlmeyer2024udfs} for enumerating small expression DAGs together with the dependence measures from the previous section to find valid substitutions that we iteratively use for reducing the dimension of symbolic regression problems. We evaluate our approach in Section~\ref{sec:experiments} on the Wikipedia eponymous equations data set \citep{bayesianscientist_guimera20} and on the Feynman symbolic regression data set\footnote{\url{https://space.mit.edu/home/tegmark/aifeynman.html}}. Finally, we draw some conclusions in Section~\ref{sec:conclusions}.

%------------------------------------------------------
% Substitutions
%------------------------------------------------------
\section{Substitutions}
\label{sec:substitutions}

Here, we formally introduce the concept of substitutions that we are going to use for reducing the complexity of symbolic regression problems. Based on the example in the introduction, the concept should be intuitively clear, we simply replace some of the variables with a function thereof. This intuition is formalized in the definition of \emph{input substitutions}.

\paragraph{Input substitutions.}

Let $\funcdef{f}{\reals^d}{\reals}$ be a function, let $I\subseteq [d]=\{1,\ldots,d\}$ with $|I| >1$ be an index set, and let $\funcdef{g}{\reals^{|I|}}{\reals^k}$ be a function with $k < |I|$. Then $g$ is an \emph{input substitution} if there exists a function $\funcdef{f_g}{\reals^{d - |I| + k}}{\reals}$ such that 
\begin{align*}
    f(x) &= f_g\big(g(x_I), x_{\setminus I}\big)
\end{align*}
for all $x\in\reals^d$. Here, $x_I$ denotes the projection of $x$ onto the indices in $I$ and $x_{\setminus I}$  denotes the projection of $x$ onto the indices in $[d]\setminus I$.

\vspace{\baselineskip}

In the context of regression problems, where we want to find the functional dependence $f$ between the input variables $x$ and the output variable $y$, we can replace the regression problem for $x$ and $y$ with a regression problem for input variables $g(x_I), x_{\setminus I}$ and output variable $y$. That is, for a given input substitution $g$, we are looking for a function $f_g$ such that
\[
f(x) = f_g\big(g(x_I), x_{\setminus I}\big).
\]
Thus, the function $f$ can be reconstructed symbolically from $g$ and $f_g$. The four substitutions that are implemented in AIFeynman~\citep{feynmanAI2_udrescu20} are special cases of bivariate input substitutions, namely, 
\[
g(x_i,x_j) = x_i \textrm{ op } x_j \,\textrm{ with }\,\textrm{op} \in \{+,-,\cdot,/\} \textrm{ and } i,j\in [d].
\]
When searching for input substitutions, another, less intuitive type of substitution makes sense, namely, substitutions that involve not only the input variables but also the output variable. We call this type of substitution an \emph{out-input substitution}.

\paragraph{Out-input substitutions.}

Let $\funcdef{f}{\reals^d}{\reals}$ be a function, $I\subseteq [d]$ with $1 \leq |I| <d$ an index set, and $\funcdef{h}{\reals^{|I|+1}}{\reals}$. Then $h$ is an \emph{out-input substitution} if there exist functions 
\[
\funcdef{g}{\reals^{d - |I|}}{\reals}
\quad\textrm{and}\quad 
\funcdef{f_g}{\reals^{|I| + 1}}{\reals}
\]
such that 
\[
h(x_I, f(x)) = g(x_{\setminus I}) \:\textrm{ and }\: f(x) = f_g\big(g(x_{\setminus I}), x_I\big) 
\]
for all $x\in\reals^d$.

\vspace{\baselineskip}

Conceptually, a \emph{simple} out-input substitution $h$ is used in an auxiliary symbolic regression problem to find a \emph{complex} input substitution $g$. The auxiliary regression problem has input variables $x_{\setminus I}$ and output variable $h(x_I, y)$, that is, we are looking for a functional dependence of the form,
\[
h(x_I, y) = g(x_{\setminus I}).
\]
Given $h$ and a solution $g$ of the auxiliary problem in symbolic form, we can solve the equation $h(x_I, y) = g(x_{\setminus I})$ symbolically for $y$. That is, $y$ becomes a symbolic expression  
\[
y = f_g\big(g(x_{\setminus I}), x_I\big),
\]
and the function $f$ that is sought in the original regression problem can be reconstructed from $g$ and $f_g$ as
\[
f(x) = f_g\big(g(x_{\setminus I}), x_I\big).
\]

The following example demonstrates that it can be beneficial to also consider out-input substitutions. The function
\[
f(x_1,x_2,x_3) = x_1x_2x_3 + \frac{x_1\big(x_2+\log(x_2)\big)}{x_3} 
\]
does not have a \emph{simple} input substitution, but the simple out-input substitution $h(x_1,y) = y/x_1$ with $I=\{1\}$. As one can easily check, the more \emph{complex} (input substitution) function
\[
g(x_2,x_3) = x_2x_3 + \frac{x_2+\log(x_2)}{x_3}
\]
and the function $\, f_g(z,x_1) = x_1 z\,$ satisfy
\begin{align*}
&h\big(x_1,f(x_1,x_2,x_3)\big)\, = g(x_2,x_3) \:\textrm{ and}\\
&f(x_1,x_2,x_3) = x_1 g(x_2,x_3) =  f_g\big(g(x_2,x_3),x_1\big). 
\end{align*}

For both substitution types, that is, input and out-input substitutions, the substituted regression problems, including the auxiliary problems, have fewer variables than the original problems, and thus should be easier to solve for symbolic regression algorithms. In Section~\ref{sec:experiments} we show that this is indeed the case for many state-of-the-art symbolic regression algorithms.

%------------------------------------------------------
% Functional Dependence Measures
%------------------------------------------------------
\section{Functional Dependence Measures}
\label{sec:funcDependence}

The fundamental assumption behind regression problems is that there exists a functional dependence between input and output variables. This assumption can be tested with functional dependence tests, or more generally by functional dependence measures. Functional dependence measures use random samples $(X_i, Y_i)_{i\in [n]}$ to quantify the degree to which a function $f(X)=Y$ between two random variables $X$ and $Y$ exists. In general, it is necessary to introduce additional constraints on the function class from which $f$ is taken. Otherwise, for pairwise distinct samples $(X_i, Y_i)_{i\in [n]}$, one always has the discrete function with $f(X_i) = Y_i$.
Classical and widely used dependence measures are the Pearson correlation coefficient~\cite{pearson:1920} for linear functions or Spearman's rank correlation~\cite{spearman:1904} for monotone functions. Recently, \citet{chatterjee:2020} introduced a new rank-based dependence measure, which works for general measurable univariate functions. Follow-up work by \citet{deb:2020} and \citet{azadkia:2021} extend this measure to multivariate functions. In the following, we describe the latter three dependence measures. 

We start with a description of \citet{chatterjee:2020}'s correlation coefficient. Given $n$ pairs of random variables $(X_i, Y_i)$ that have been drawn i.i.d.\ from some probability distribution and the variables $X_i$ have been sorted in ascending order. The Chatterjee coefficient is then defined as 
\[
\xi_n(X,Y) = 1 - \frac{n}{2}\frac{\sum_{i=1}^{n-1}|r_{i+1} - r_i|}{\sum_{i=1}^n l_i(n-l_i)}\,,
\]
where $r_i$ denotes the rank of $Y_i$, that is, the number of indices $j$ such that $Y_j \leq Y_i$. Analogously, $l_i$ is defined as the number of indices $j$ such that $Y_j \geq Y_i$.
For $n\to\infty$, this coefficient $\xi_n(X,Y)$ converges to $0$ if $X_i$ and $Y_i$ are independent variables. It converges to $1$ if there exists a measurable function $\funcdef{f}{\reals}{\reals}$ such that $Y = f(X)$.

This can be explained intuitively as follows. If all $X_i$ and all $Y_i$ are distinct, then, after a small calculation, the coefficient simplifies to 
\[
\xi_n(X,Y) = 1 - \frac{3\sum_{i=1}^{n-1}|r_{i+1} - r_i|}{n^2 - 1}\,.
\]
Moreover, if we assume further that $Y_i = f(X_i)$ for a continuous function $f$ and that the $X_i$ are sorted and densely sampled, then not only
\[
\big|Y_{i+1} - Y_i\big| = \big|f(X_{i+1})-f(X_i)\big|
\]
should be small, but also the rank difference $|r_{i+1} - r_i|$. In this case, we have, 
\[
\sum_{i=1}^{n-1} |r_{i+1} - r_i| \in O(n),
\]
which means that $\xi_n(X,Y)$ converges to $1$.

The Chatterjee coefficient is defined only for univariate functions $\funcdef{f}{\reals}{\reals}$ and thus is not suitable for our intended application in symbolic regression where we need to decide functional dependence of multivariate functions. 
\citet{azadkia:2021} have generalized the Chatterjee coefficient to the multivariate case. Their coefficient, called $\codec$, is designed for detecting a functional dependence between random variables, conditioned on another set of variables. In our application, we can simply condition on the empty set. $\codec$ is defined for random variables $X$ that take values in $\reals^d$ and $Y$ that take values in $\reals$. Let $(X_i, Y_i)_{i\in [n]}$ be a sample that is drawn i.i.d from some probability distribution on $\reals^d\times \reals$. The values $r_i$ and $l_i$ are the upper respectively lower ranks as for the Chatterjee coefficient. Let $\nu (i)$ denote the index such that, $X_{\nu (i)}$ is the nearest neighbor of $X_i$, where ties are broken arbitrarily.  $\codec$ is now defined as
\begin{align*}
    \codec_n(X, Y) = \frac{\sum_{i=1}^n \big(n \min\{ r_i, r_{\nu (i)}\} - l_i^2\big)}{\sum_{i=1}^n l_i(n-l_i)}\,.
\end{align*}
As for the Chatterjee coefficient, if $X$ and $Y$ are independent, then $\codec_n(X, Y)$ converges to $0$ for $n\to\infty$. It converges to $1$ if $Y$ is a measurable function of $X$. The intuition behind $\codec$ is similar to the intuition behind the Chatterjee coefficient as well. Using the well known identity 
\[
\min\{a,b\} = \frac{1}{2}(a + b - |a - b|),
\] 
we can rewrite the numerator of the $\codec$ coefficient as
\begin{align*} 
    &\sum_{i=1}^n \Big( n\min\{r_i, r_{\nu (i)}\} - l_i^2\Big) \\
    %&\qquad\qquad\quad= \sum_{i=1}^n n\min\{r_i, r_{\nu (i)}\} - \sum_{i=1}^n l_i^2 \\
    &\qquad\qquad\quad= \frac{n}{2}\Big(R + S -  \sum_{i=1}^n |r_i - r_{\nu (i)}|\Big) - L, 
\end{align*} 
where 
\[
L = \sum_{i=1}^n l_i^2,\, R = \sum_{i=1}^n r_i, \,\textrm{ and }\, S = \sum_{i=1}^n r_{\nu (i)}.
\]
The $\codec$ coefficient now becomes,
\begin{align*}
&\codec_n(X, Y) = \\
&\qquad\qquad\qquad\frac{\frac{n}{2}\left(R + S -  \sum_{i=1}^n |r_i - r_{\nu (i)}|\right) - L}{\sum_{i=1}^n l_i(n-l_i)}\,,
\end{align*}
which becomes large, when $\sum_{i=1}^n |r_i - r_{\nu (i)}|$ is small. This happens exactly when the $Y_i$ values, and thus the ranks, of sample neighboring points are close.  

Both, the Chatterjee and the $\codec$ coefficient are special cases of a more general class of dependence measures, called $\kmac$, short for kernel measure of association~\citep{deb:2020}. $\kmac$ measures also quantify how close the $Y_i$ values are for sample points $X_i$ that are close to each other. To do so, $\kmac$ measures use geometric graphs, such as,  $k$-nearest-neighbor graphs or a Euclidean minimum spanning trees, on the sample points $X_i$ to encode which sample points should be considered close to each other. The distance between the values $Y_i$ is measured through a mapping into a reproducing kernel Hilbert space. In theory, this allows us to assess functional dependence not only on $\reals^d$ but on arbitrary topological spaces, as long as we have a notion of closeness for the sample points.

%------------------------------------------------------
% Finding and Using Substitutions
%------------------------------------------------------
\section{Finding and Using Substitutions}
\label{sec:loop}

Our goal is to find input substitutions and out-input substitutions to transform the observations
\[
\big(x^{(i)},y^{(i)}\big) \in \reals^d\times \reals \quad\textrm{for}\, i\in [n],
\]
that define a symbolic regression problem into a simpler regression problem with fewer variables. 

\paragraph{Input substitutions.} 

Let $I\subseteq [d]$ with $|I| >1$ be an index set, and let $\funcdef{g}{\reals^{|I|}}{\reals^k}$ be a function with $k < |I|$ that transforms the observations into 
\[
\Big(\big(g(x^{(i)}_I), x_{\setminus I}\big), y^{(i)} \Big) \in \reals^{d - |I| + k}\times \reals.
\]
If there is a functional dependence between the transformed input variables $\big(g(x^{(i)}_I), x_{\setminus I}\big)$ and the output variables $y^{(i)}$, then the transformed observations amount to an \emph{auxiliary} symbolic regression problem with only $d - |I| + k$ input variables. From $g$ and a symbolic solution $f_g$ of the auxiliary regression problem, a solution $f$ of the original problem can be reconstructed.

\paragraph{Out-input substitutions.}

Let $I\subseteq [d]$ with $1 \leq |I| <d$ be an index set, and let $\funcdef{h}{\reals^{|I|+1}}{\reals}$ be a function that transforms the observations into 
\[
\Big(x_{\setminus I}^{(i)},h\big(x_I^{(i)},y^{(i)}\big)\Big) \in \reals^{d -|I|}\times \reals.
\]
If there is a functional dependence between the transformed input variables $x_{\setminus I}^{(i)}$ and the output variable $h\big(x_I^{(i)},y^{(i)}\big)$, then the transformed observations amount to a symbolic regression problem with only $d - |I|$ input variables. Assume that there is such a dependence and that the entailed \emph{auxiliary} symbolic regression problem is solved by a function $\funcdef{g}{\reals^{d - |I|}}{\reals}$. We can use the symbolically given functions $h$ and $g$ to derive a symbolic function $f_g$ by solving the equation $h(x_I, y) = g(x_{\setminus I})$ for $y$. Examples can be found in Section~\ref{sec:substitutions} and on an example in the supplement.%Section~\ref{sec:substitutions} and in the full version of the paper.
The symbolic solution $f$ of the original symbolic regression problem can be reconstructed from $g$ and $f_g$. As mentioned before, $g$ is actually an input substitution that might be too complex to be found directly.

\vspace{\baselineskip}

Given the observations that define a symbolic regression problem, we can use both input substitutions $g$ and in-output substitutions $h$ to derive a simpler, auxiliary symbolic regression problem. Therefore, the task becomes to find valid substitutions $g$ and $h$. We address this task by systematically searching for $g$ and $h$ in a space of functions with succinct symbolic representations. In their search-based regressor, \citet{kahlmeyer2024udfs} enumerate expression DAGs up to a given size. For a fixed budget of operator nodes, the representation by DAGs allows covering more functions than the commonly used expression tree representation, because in DAGs common subexpressions need to be represented only once. Here, we use the enumeration of small expression DAGs to systematically enumerate \emph{simple} candidate substitutions $g$ and $h$. By a functional dependence test on transformed observations, we can decide if a candidate is indeed a valid substitution. Substitutions can be used not only on the observations for the original symbolic regression problem, but also for auxiliary problems that are given by transformed observations. That is, we can continue and iteratively simplify the problems further. 

\paragraph{Beam search for symbolic regression.}

The iterative search for substitutions is naturally organized in a \emph{search-tree}. At the root of the search tree is the original regression problem, which is given by the observations. The children of any node in the tree are auxiliary regression problems that are derived from valid substitutions, and the leafs of the tree are regression problems for which no substitution can be found. By the definition of substitutions, regression problems at child nodes always have fewer variable nodes than the regression problems at their parent nodes. The regression problems at all nodes of the search tree can be addressed by any symbolic regression algorithm. 

\begin{figure}[ht]
    \centering
    \tikzset{every picture/.style={line width=0.75pt}} %set default line width to 0.75pt        

\begin{tikzpicture}[x=0.75pt,y=0.75pt,yscale=-1,xscale=1]
%uncomment if require: \path (0,465); %set diagram left start at 0, and has height of 465

%Rounded Rect [id:dp44592909797100744] 
\draw  [fill={rgb, 255:red, 255; green, 255; blue, 255 }  ,fill opacity=1 ][general shadow={fill={rgb, 255:red, 155; green, 155; blue, 155 }  ,shadow xshift=1.5pt,shadow yshift=-1.5pt, opacity=1 }] (150,197.11) .. controls (150,193.18) and (153.18,190) .. (157.11,190) -- (262.89,190) .. controls (266.82,190) and (270,193.18) .. (270,197.11) -- (270,252.89) .. controls (270,256.82) and (266.82,260) .. (262.89,260) -- (157.11,260) .. controls (153.18,260) and (150,256.82) .. (150,252.89) -- cycle ;
%Straight Lines [id:da7517503808631956] 
\draw    (190,180) -- (230,180) -- (230,278) ;
\draw [shift={(230,280)}, rotate = 270] [fill={rgb, 255:red, 0; green, 0; blue, 0 }  ][line width=0.08]  [draw opacity=0] (12,-3) -- (0,0) -- (12,3) -- cycle    ;
%Rounded Rect [id:dp8785093639217305] 
\draw  [fill={rgb, 255:red, 255; green, 255; blue, 255 }  ,fill opacity=1 ][general shadow={fill={rgb, 255:red, 155; green, 155; blue, 155 }  ,shadow xshift=1.5pt,shadow yshift=-1.5pt, opacity=1 }] (10,197.11) .. controls (10,193.18) and (13.18,190) .. (17.11,190) -- (122.89,190) .. controls (126.82,190) and (130,193.18) .. (130,197.11) -- (130,252.89) .. controls (130,256.82) and (126.82,260) .. (122.89,260) -- (17.11,260) .. controls (13.18,260) and (10,256.82) .. (10,252.89) -- cycle ;
%Rounded Rect [id:dp987750443874246] 
\draw  [fill={rgb, 255:red, 255; green, 255; blue, 255 }  ,fill opacity=1 ][general shadow={fill={rgb, 255:red, 155; green, 155; blue, 155 }  ,shadow xshift=1.5pt,shadow yshift=-1.5pt, opacity=1 }] (11.87,37.44) .. controls (11.87,32.78) and (15.65,29) .. (20.31,29) -- (261.56,29) .. controls (266.22,29) and (270,32.78) .. (270,37.44) -- (270,161.56) .. controls (270,166.22) and (266.22,170) .. (261.56,170) -- (20.31,170) .. controls (15.65,170) and (11.87,166.22) .. (11.87,161.56) -- cycle ;
%Shape: Polygon Curved [id:ds14705833767534437] 
\draw  [fill={rgb, 255:red, 230; green, 230; blue, 230 }  ,fill opacity=1 ] (270,40) .. controls (270.45,45.58) and (269.83,92.83) .. (270,100) .. controls (270.17,107.17) and (42.71,93.83) .. (38.87,66) .. controls (35.04,38.17) and (269.55,34.42) .. (270,40) -- cycle ;
%Shape: Polygon Curved [id:ds3917754411577893] 
\draw  [fill={rgb, 255:red, 206; green, 206; blue, 206 }  ,fill opacity=1 ] (270,50) .. controls (270.83,53.83) and (270.17,83.5) .. (270,90) .. controls (269.83,96.5) and (104.83,83.83) .. (135,65) .. controls (165.17,46.17) and (269.17,46.17) .. (270,50) -- cycle ;
%Straight Lines [id:da4156374237086309] 
\draw  [dash pattern={on 0.84pt off 2.51pt}]  (36.87,68) -- (37,120) ;
%Straight Lines [id:da5221875962629928] 
\draw    (36.87,68) -- (117,65.11) ;
\draw [shift={(120,65)}, rotate = 177.93] [fill={rgb, 255:red, 0; green, 0; blue, 0 }  ][line width=0.08]  [draw opacity=0] (8.93,-4.29) -- (0,0) -- (8.93,4.29) -- cycle    ;
%Straight Lines [id:da8097896373233685] 
\draw    (135,65) -- (158.17,65) -- (217,65) ;
\draw [shift={(220,65)}, rotate = 180] [fill={rgb, 255:red, 0; green, 0; blue, 0 }  ][line width=0.08]  [draw opacity=0] (8.93,-4.29) -- (0,0) -- (8.93,4.29) -- cycle    ;
%Shape: Rectangle [id:dp3642081895547761] 
\draw   (20,120) -- (80,120) -- (80,150) -- (20,150) -- cycle ;
%Shape: Rectangle [id:dp6300763342914876] 
\draw   (90,120) -- (215,120) -- (215,150) -- (90,150) -- cycle ;
%Shape: Rectangle [id:dp5310017459507214] 
\draw   (230,120) -- (255,120) -- (255,150) -- (230,150) -- cycle ;
%Straight Lines [id:da7816420241594317] 
\draw  [dash pattern={on 0.84pt off 2.51pt}]  (135,65) -- (135,120) ;
%Straight Lines [id:da7100424052721732] 
\draw  [dash pattern={on 0.84pt off 2.51pt}]  (236,65) -- (236,120) ;
%Straight Lines [id:da4054523300809928] 
\draw    (36.87,21) -- (36.87,49) ;
\draw [shift={(36.87,51)}, rotate = 270] [fill={rgb, 255:red, 0; green, 0; blue, 0 }  ][line width=0.08]  [draw opacity=0] (12,-3) -- (0,0) -- (12,3) -- cycle    ;
%Shape: Polygon Curved [id:ds8771944596266505] 
\draw  [fill={rgb, 255:red, 175; green, 175; blue, 175 }  ,fill opacity=1 ] (270,50) .. controls (270.17,49.17) and (270.17,79.17) .. (270,80) .. controls (269.83,80.83) and (238.83,80.83) .. (235,65) .. controls (231.17,49.17) and (269.83,50.83) .. (270,50) -- cycle ;
%Straight Lines [id:da6242179135781518] 
\draw    (145,150) -- (145,160) ;
%Shape: Rectangle [id:dp8843692800528372] 
\draw  [fill={rgb, 255:red, 255; green, 255; blue, 255 }  ,fill opacity=1 ][general shadow={fill={rgb, 255:red, 155; green, 155; blue, 155 }  ,shadow xshift=1.5pt,shadow yshift=-1.5pt, opacity=1 }] (30,280) -- (250,280) -- (250,300) -- (30,300) -- cycle ;
%Shape: Circle [id:dp6345850727141812] 
\draw  [fill={rgb, 255:red, 255; green, 255; blue, 255 }  ,fill opacity=1 ] (120,65) .. controls (120,56.72) and (126.72,50) .. (135,50) .. controls (143.28,50) and (150,56.72) .. (150,65) .. controls (150,73.28) and (143.28,80) .. (135,80) .. controls (126.72,80) and (120,73.28) .. (120,65) -- cycle ;
%Shape: Circle [id:dp8446432167287339] 
\draw  [fill={rgb, 255:red, 255; green, 255; blue, 255 }  ,fill opacity=1 ] (23.87,66) .. controls (23.87,57.72) and (30.59,51) .. (38.87,51) .. controls (47.16,51) and (53.87,57.72) .. (53.87,66) .. controls (53.87,74.28) and (47.16,81) .. (38.87,81) .. controls (30.59,81) and (23.87,74.28) .. (23.87,66) -- cycle ;
%Shape: Circle [id:dp615943379944182] 
\draw  [fill={rgb, 255:red, 255; green, 255; blue, 255 }  ,fill opacity=1 ] (220,65) .. controls (220,56.72) and (226.72,50) .. (235,50) .. controls (243.28,50) and (250,56.72) .. (250,65) .. controls (250,73.28) and (243.28,80) .. (235,80) .. controls (226.72,80) and (220,73.28) .. (220,65) -- cycle ;
%Straight Lines [id:da07441466464499924] 
\draw    (50,150) -- (50,160) -- (240,160) -- (240,150) ;
%Straight Lines [id:da26088190465696504] 
\draw    (90,230) -- (90,278) ;
\draw [shift={(90,280)}, rotate = 270] [fill={rgb, 255:red, 0; green, 0; blue, 0 }  ][line width=0.08]  [draw opacity=0] (12,-3) -- (0,0) -- (12,3) -- cycle    ;
%Straight Lines [id:da14931623785356296] 
\draw    (190,230) -- (190,278) ;
\draw [shift={(190,280)}, rotate = 270] [fill={rgb, 255:red, 0; green, 0; blue, 0 }  ][line width=0.08]  [draw opacity=0] (12,-3) -- (0,0) -- (12,3) -- cycle    ;
%Shape: Rectangle [id:dp9247955844037616] 
\draw   (18.11,234.5) -- (68.11,234.5) -- (68.11,255) -- (18.11,255) -- cycle ;
%Straight Lines [id:da4200510546535313] 
\draw    (50,255) -- (50,278) ;
\draw [shift={(50,280)}, rotate = 270] [fill={rgb, 255:red, 0; green, 0; blue, 0 }  ][line width=0.08]  [draw opacity=0] (12,-3) -- (0,0) -- (12,3) -- cycle    ;
%Rounded Rect [id:dp4739151688071793] 
\draw  [fill={rgb, 255:red, 255; green, 255; blue, 255 }  ,fill opacity=1 ][general shadow={fill={rgb, 255:red, 155; green, 155; blue, 155 }  ,shadow xshift=1.5pt,shadow yshift=-1.5pt, opacity=1 }] (60,219.4) .. controls (60,214.21) and (64.21,210) .. (69.4,210) -- (210.6,210) .. controls (215.79,210) and (220,214.21) .. (220,219.4) -- (220,221.1) .. controls (220,226.29) and (215.79,230.5) .. (210.6,230.5) -- (69.4,230.5) .. controls (64.21,230.5) and (60,226.29) .. (60,221.1) -- cycle ;
%Straight Lines [id:da6581111257667738] 
\draw    (90,247) -- (69,247) ;
\draw [shift={(67,247)}, rotate = 360] [fill={rgb, 255:red, 0; green, 0; blue, 0 }  ][line width=0.08]  [draw opacity=0] (12,-3) -- (0,0) -- (12,3) -- cycle    ;
%Straight Lines [id:da24164906914122442] 
\draw    (190,160) -- (190,208) ;
\draw [shift={(190,210)}, rotate = 270] [fill={rgb, 255:red, 0; green, 0; blue, 0 }  ][line width=0.08]  [draw opacity=0] (12,-3) -- (0,0) -- (12,3) -- cycle    ;
%Straight Lines [id:da4309138976361455] 
\draw    (90,160) -- (90,208) ;
\draw [shift={(90,210)}, rotate = 270] [fill={rgb, 255:red, 0; green, 0; blue, 0 }  ][line width=0.08]  [draw opacity=0] (12,-3) -- (0,0) -- (12,3) -- cycle    ;
%Straight Lines [id:da786983944508979] 
\draw    (90,194) -- (50,194) -- (50,232) ;
\draw [shift={(50,234)}, rotate = 270] [fill={rgb, 255:red, 0; green, 0; blue, 0 }  ][line width=0.08]  [draw opacity=0] (12,-3) -- (0,0) -- (12,3) -- cycle    ;
%Shape: Rectangle [id:dp9307041469817767] 
\draw  [draw opacity=0][fill={rgb, 255:red, 255; green, 255; blue, 255 }  ,fill opacity=0.75 ][line width=0.75]  (45,191.5) -- (57.17,191.5) -- (57.17,211) -- (45,211) -- cycle ;
%Shape: Rectangle [id:dp16914072892187337] 
\draw  [draw opacity=0][fill={rgb, 255:red, 255; green, 255; blue, 255 }  ,fill opacity=0.75 ][line width=0.75]  (57,192.5) -- (69,192.5) -- (69,201) -- (57,201) -- cycle ;

% Text Node
\draw (38.87,66) node    {$0.85$};
% Text Node
\draw (135,65) node    {$0.9$};
% Text Node
\draw (235,65) node    {$0.92$};
% Text Node
\draw (50,135) node    {$x^{( i)} ,\ y^{( i)}$};
% Text Node
\draw (152.5,135) node    {$( g( x_{I}) ,x_{\setminus I})^{( i)} ,\ y^{( i)}$};
% Text Node
\draw (242.5,135) node    {$\dotsc $};
% Text Node
\draw (140,220.25) node   [align=left] {symbolic regressor};
% Text Node
\draw (140,290) node    {$f( x) =y$};
% Text Node
\draw (36.37,11) node    {$x^{( i)} ,\ y^{( i)}$};
% Text Node
\draw (234,191) node [anchor=north west][inner sep=0.75pt]   [align=left] {input};
% Text Node
\draw (11,192) node [anchor=north west][inner sep=0.75pt]   [align=left] {out-input};
% Text Node
\draw (171,240.4) node [anchor=north west][inner sep=0.75pt]    {$f_{g}$};
% Text Node
\draw (91,242.4) node [anchor=north west][inner sep=0.75pt]    {$g$};
% Text Node
\draw (42.11,245.75) node   [align=left] {CAS};
% Text Node
\draw (52,262.4) node [anchor=north west][inner sep=0.75pt]    {$f_{g}$};
% Text Node
\draw (177,175.4) node [anchor=north west][inner sep=0.75pt]    {$g$};
% Text Node
\draw (91,172.4) node [anchor=north west][inner sep=0.75pt]    {$h$};

\end{tikzpicture}
    \caption{Illustration of the beam search approach for symbolic regression. Each node of the beam search tree has an associated regression data set and a functional dependence score for this data set. The edges are labeled by candidate substitutions. The highest scoring regression problems, over all levels, are passed to a symbolic regression algorithm. For input substitutions $g$, a symbolic regression algorithms computes a function $f_g$. For out-input substitutions $h$, the symbolic regression algorithm computes $g$, which is passed together with $h$ to a computer algebra system (CAS) to find a function $f_g$. A solution $f$ of the original symbolic regression problem is reconstructed from all functions $g$ and $f_g$ along the path from the root to the overall highest scoring node in the beam search tree.
    }
    \label{fig:flowchart}
\end{figure}
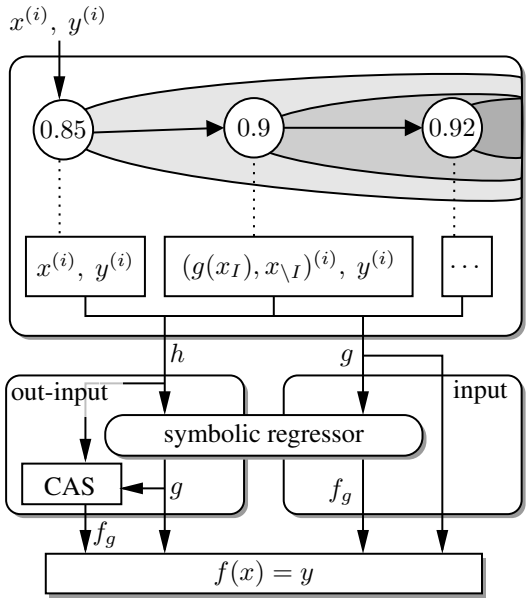

However, the functional dependence measures from Section~\ref{sec:funcDependence} do not provide a definite answer, but only a functional dependence score between $0$ and $1$, where scores close to $1$ indicate functional dependence and scores close to $0$ its absence. The scores can be used to selectively explore the search tree by a \emph{beam search}. That is, every candidate substitution becomes a node in the search tree, but now the nodes are weighted by a score. As usual in beam search, a beam size hyperparameter controls the size of the search space by limiting the number of nodes on each level to the beam-size-many highest scoring substitutions. The search tree is then explored in a breadth-first manner, where the beam size parameter restricts the number of paths that are followed. After the search has ended, the regression problems on the path that is leading to the highest scoring regression problem are passed on to an established symbolic regression algorithm. The beam-search algorithm for a symbolic regression is illustrated in Figure~\ref{fig:flowchart}.

%------------------------------------------------------
% Experiments
%------------------------------------------------------
\section{Experiments}
\label{sec:experiments}

Since our work is motivated by the observation that symbolic regression becomes easier for smaller expressions, we evaluate the ability of our beam-search approach to reduce the complexity of symbolic regression problems in the following experiments. We assess the dimension-reduction ability by the \emph{reduction rate} for formulas with known symbolic expressions. Our beam search is parameterized by a choice of functional dependence measure, by the beam size, by the choice of substitution type, by the symbolic regression algorithm, and by the expression DAG search space. To gauge the effect of the first three parameters, we evaluate the reduction rate for different dependence measures, different beam sizes, and three choices for the substitution types. In a second experiment, we evaluate the effectiveness of combining our beam search with different state-of-the-art symbolic regression algorithms. In this experiment, we compare the \emph{recovery rate} of the symbolic regression algorithms with and without our beam search. For efficiency reasons, we keep the DAG search space small and fixed in all experiments.

\paragraph{Datasets.}

For our experiments, we have used two sets of regression problems, namely, Wikipedia’s list of 880 eponymous equations~\citep{bayesianscientist_guimera20} and 114 formulas that were extracted from the Feynman lecture notes of physics~\citep{feynmanAI_udrescu20}. We provide more details about both data sets in the supplement.

\paragraph{Implementation Details.}

At each node of the beam search tree, we search for substitutions in a space of expression DAGs. To keep the search space small, we consider only expression DAGs with at most one intermediary node and one output node, that is, scalar functions with $k=1$. 

The $\kmac$ measures are parameterized by a geometric graph and a kernel function. Here, we follow the original paper by \citet{deb:2020}, and use a one-nearest neighbor graph and a Gaussian RBF kernel.
%In our case, we used a standard Gaussian kernel $K(x, y) = \exp\left(-\gamma||x - y||^2\right)$, where we set $\gamma = 1$.

For out-input substitutions, we derive $f_g$ from $h$ and $g$ by using the computer algebra system SymPy~\citep{sympy_meurer17}.

\paragraph{Hardware.}

All experiments were run on a computer with an Intel Xeon Gold 6226R 64-core processor, 128 GB of RAM, running Python 3.10.

%------------------------------------------------------
% Parameter Study
%------------------------------------------------------
\subsection{Parameter Study}

For evaluating the effect of the choice of dependence measure, the beam size parameter, and the substitution type, we first need to define the reduction rate for formulas with known symbolic expressions, before we compute these rates in different experimental settings. 

\paragraph{Reduction rate.}

For computing the reduction rate in a given parameter setting, we run our beam search on formulas with known symbolic expressions as described in Section~\ref{sec:loop}, except for the last step. That is, we do not pass the reduced regression problems to a symbolic regression algorithm. Instead, we check the correctness of the selected substitutions. The \emph{reduction rate} is the best dimension reduction that is achieved by the beam search, that is, 
\[
\textrm{reduction rate }  =\, 1- \frac{\min_{\textrm{nodes}}\, \# \textrm{vars of reduced problem}}{\# \textrm{vars of original problem}}.
\]

We use SymPy for checking the validity of a substitution. Let $g$ be a candidate input substitution that replaces the input variables $x_I$ with $g(x_I)$. Since the candidate substitutions are given symbolically, we can introduce a new variable symbol $\gamma$ and solve the equation $\gamma = g(x_I)$ for some input variable $x_i$ with $i
\in I$. Then, we plug the expression for $x_i$, which depends on $\gamma$, into the known ground truth formula for $f$. This gives us a symbolic expression $\hat f$. If $g$ is a valid substitution, then all input variables $x_j, j\in I\setminus \{i\}$ can be eliminated from $\hat f$. That is, $\hat f$ has an equivalent symbolic representation that only depends on $\gamma$ and the input variables $x_{\setminus I}$. The validity of a candidate out-input substitution $h$ that replaces the input variables $x_I$ and the output variable $y$ with $h(x_I,y)$ can be checked similarly. Given the ground truth formula $f$ in symbolic form, we can substitute $f(x)$ for $y$ in $h(x_I,y)$. If $h$ is valid, then the symbolic expression for $h(x_I,f(x))$ does not depend on $x_I$. We provide examples in the supplement.  

\paragraph{Choice of dependence measure.}

We measure the reduction rate of our beam search when used with the $\codec$ and $\kmac$ dependence measures, and a simple baseline measure. The baseline measure is a simple volume heuristic. The observations $(x^{(i)},y^{(i)})_{i\in[n]}$ are contained in a submanifold of $\reals^{d+1}$ if they are sampled from a well-behaved function~$f$ with $y=f(x)$. In our volume heuristics, we compute for every point $x_{(i)}$ the volume of the parallelepiped that is spanned by the difference vectors of $x^{(i)}$ to its $d+1$ nearest neighbors. These volumes should be close to zero if the observations are sampled densely from a well-behaved function. The volume is given by the determinant of the matrix of difference vectors. For the dependence measure, we take the average over the $n$ volumes, 

For the experiment, we follow the experimental protocol by \citet{srbench_lacava21} and sample the ground-truth functions in the eponymous equations and the Feynman equations data sets, for which the true functional dependencies $y = f(x)$ are known. The samples for Feynman equations data set are directly given by \citet{srbench_lacava21}, who also add different levels of Gaussian noise to the sample points. Sampling from the functions in the eponymous equations data set is described in the supplement. 

\begin{figure}[h!]
    \centering
    \includegraphics[width = \columnwidth]{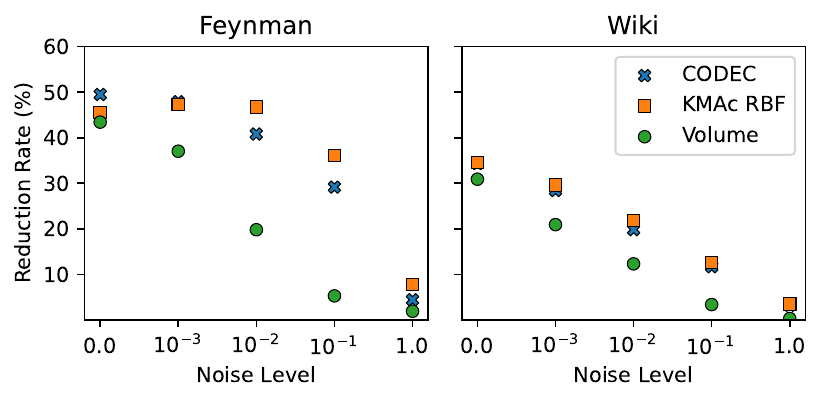}
    \caption{Average reduction rates of our beam search, with beam size $1$, on the eponymous equations (Wiki) and the Feynman equations (Feynman) data sets for the functional dependence measures $\codec$ and $\kmac$, and a volume-based baseline measure.}
    \label{fig:dim_reduction}
\end{figure}

We report average reduction rates for the different dependence measures and beam size $1$ in Figure~\ref{fig:dim_reduction}. In the noise-free case, all three dependence measures perform similarly. The number of input variables is reduced by $\sim$50\% for the Feynman equations data set and by $\sim$35\% for the eponymous equations data set. At non-zero noise levels, however, the baseline measure breaks down, whereas $\codec$ and $\kmac$ still achieve substantial average reduction rates.

While $\codec$ and $\kmac$ perform similarly in terms of average reduction rate, they differ significantly, almost an order of magnitude, in running times, as can be seen in Table~\ref{tab:runtime}. Therefore, we use our beam search with $\codec$ in the following.
\begin{table}[h!]
    \centering
    \begin{tabular}{llll}
    & $\codec$ & $\kmac$ & Volume\\
    \toprule
    Feynman&$3.32\pm0.08$&$24.29\pm0.29$&$21.58\pm0.44$\\
    Wiki&$2.88\pm0.02$&$23.24\pm0.14$&$20.92\pm0.15$\\
    \end{tabular}
    \caption{Average running times (in milliseconds) for the identification of a single substitution on the eponymous equations and the Feynman equations data sets. The average and standard deviations have been computed over 50 runs.}
    \label{tab:runtime}
\end{table}

\paragraph{Beam size.}

We measure the reduction rate of our beam search for five values of the beam size parameter, while keeping the functional dependence measure $\codec$ fixed. A larger beam size broadens the view of the search process and potentially finds better substitutions, however, at the cost of increased running times. We report results for beam sizes $1$ to $5$ in Table~\ref{tab:beamsize}. Since the reduction rate improves only marginally with increasing beam size, we use our beam search with beam size $1$ in the following.

\begin{table}[ht]
    \centering
    \begin{tabular}{lccccc}
    beam size&1&2&3&4&5\\
    \toprule
    Feynman&0.489&0.504&0.491&0.491&0.491\\
    Wiki&0.346&0.354&0.354&0.357&0.358\\
    \end{tabular}
    \caption{Average reduction rates of for beam sizes $1$ to $5$.}
    \label{tab:beamsize}
\end{table}

\paragraph{Substitution type.}

So far, we have used input substitutions as well as out-input substitutions in the experiment. Here, we assess how much out-input substitutions add to the reduction-rate performance of the beam search. In Table~\ref{tab:ablation} we compare the reduction rate of our beam search for out-input and input substitutions, only input substitutions, and the four substitutions from the AIFeynman project \citep{feynmanAI2_udrescu20}. Remember that, conceptually, out-input substitutions are used for indirectly finding input substitutions that are more complex than the functions in the direct search space. Here, the search space is not very large, since we only consider expression DAGs with one input and one output node. However, as can be seen on the eponymous equations data set (Wiki), it covers more input substitutions than AIFeynman. Adding out-input substitutions increases the reduction rates significantly on both data sets.   

\begin{table}[h!]
    \centering
    \begin{tabular}{lccc}
    &out+input&input&AIFeynman\\
    \toprule
    Feynman&0.49&0.39&0.39\\
    Wiki&0.34&0.19&0.08\\
    \end{tabular}
    \caption{Comparison of the average reduction rate for out-input and input substitutions (out+input), only input substitutions, and the AIFeynman substitutions.}
    \label{tab:ablation}
\end{table}

%------------------------------------------------------
% Recovery Rate
%------------------------------------------------------
\subsection{Recovery Rate}

As described in Section~\ref{sec:loop}, our beam search can be turned into a symbolic regressor by passing reduced problem instances that have been discovered in the search process to a symbolic regression algorithm. An established measure for assessing the quality of a symbolic regression algorithm is the \emph{recovery rate}, that is, the fraction of formulas with known symbolic expressions that are recovered by a symbolic regression algorithm up to symbolic equivalence \citep{srbench_lacava21}. Here, we use again SymPy to automatically decide the symbolic equivalence of expressions. Following the symbolic regression benchmark paper by \citet{srbench_lacava21}, we also measure model fit and model complexity, besides the recovery rate. A good symbolic regression algorithm computes models that provide high recovery, have good model fits, and are of low complexity. Here, model fit is measured by the normalized root-mean-square error (NRMSE) on hold out data, and the complexity of an expression is measured by the number of nodes in a corresponding expression tree that has been simplified by SymPy.

\paragraph{Symbolic regression algorithms.}

In the experiment, we compare the recovery rate of several state-of-the-art symbolic regression algorithms in combination with our beam search, or without. The beam search in the experiment uses beam size $1$ and the $\codec$ functional dependence measure. One can roughly distinguish three types of symbolic regression algorithms, namely, genetic programming algorithms, sampling algorithms, and search algorithms. We provide a review of the different types of symbolic regression algorithms, together with examples in the supplement. For our experiment, we used our beam search with representatives for all three types: \textsc{Operon} \citep{operon_burlacu} and \textsc{PySR} \citep{cranmer2023pysr} for genetic programming, the sequence-to-sequence learning approach \textsc{Tranf} by \citet{transformer_kamienny22} and deep symbolic regression (\textsc{DSR}) by \citet{dsr2_petersen21} as two examples for sampling algorithms, and the unbiased DAG frame search (\textsc{UDFS}) by \citet{kahlmeyer2024udfs} as an example for a search algorithm. Moreover, we use the beam search also with a standard regression model, namely, polynomial regression with polynomials of degree at most two (\textsc{Poly}). 
If applicable, we set the hyperparameters of the symbolic regression algorithms so that our experiments run in reasonable time. An overview of the respective hyperparameters can be found in the supplement.

\paragraph{Results.}

Results on the Feynman equations data set are shown in Table~\ref{tab:regression_boost}. More details and results for the eponymous equations data set are provided in the supplement. Notably, the recovery rate improves significantly for all regression algorithms included in the experiment. As can be seen, this typically also affects the average model fit, because correctly recovered expressions have RMSEs of zero, or numerically very close to zero, on test data. The complexity of the expressions that are returned by the symbolic regression algorithms is not affected much by the beam search. The only exception is polynomial regression, where the expression trees shrink on average by more than 20 nodes when using the beam search.

\begin{table}[ht]
    \centering
    \begin{tabular}{lllllll}
    &\multicolumn{2}{c}{recovery}&\multicolumn{2}{c}{NRMSE}&\multicolumn{2}{c}{complexity}\\
    &base&beam&base&beam&base&beam\\
    \toprule
    \textsc{UDFS}&0.58&\textbf{0.69}&0.08&\textbf{0.03}&\textbf{10.26}&12.71\\
    DSR&0.37&\textbf{0.66}&0.14&\textbf{0.03}&\textbf{19.09}&21.17\\
    \textsc{Transf}&0.01&\textbf{0.11}&0.09&\textbf{0.03}&33.53&\textbf{28.97}\\
    \textsc{PySR}&0.51&\textbf{0.62}&0.09&\textbf{0.03}&\textbf{11.74}&14.86\\
    \textsc{Operon}&0.41&\textbf{0.52}&0.05&\textbf{0.02}&\textbf{34.81}&40.93\\
    \textsc{Poly}&0.06&\textbf{0.34}&0.18&\textbf{0.08}&54.2&\textbf{32.09}\\
    \end{tabular}
    \caption{Performance measures of different symbolic regression algorithms in combination with our search, or without (\textbf{bold} is better). Recovery rate, average normalized root-mean-square error on test data, and average model complexity on the Feynman equations data set.}
    \label{tab:regression_boost}
\end{table}

It turns out, that even for problems that are not recovered by the algorithms, the returned expressions have more subexpressions in common with the ground-truth formula when used together with our beam search. See the supplement for details.

%------------------------------------------------------
% Conclusions
%------------------------------------------------------
\section{Conclusions}
\label{sec:conclusions}

Motivated by the straightforward, yet seemingly so far mostly overlooked observation that symbolic regression becomes more difficult for more complex formulas, we have designed and implemented a dimension reduction method for symbolic regression. Given a symbolic regression problem in the form of observations, our method systematically searches for small substitutions that transform the observed data into a lower dimensional data set, that is, a data set where each observation is replaced with a data point with fewer variables. The method can be applied iteratively to the dimension-reduced data sets. We have implemented the idea of an iterative dimension reduction in a beam search. Combining the beam search with state-of-the-art symbolic regression algorithms significantly boosts the symbolic recovery performance of these algorithms.

\section*{Acknowledgements}
This work was supported by the Carl Zeiss Stiftung within the project "Interactive Inference".
 
\bibliography{aaai25}

\begin{thebibliography}{27}
\providecommand{\natexlab}[1]{#1}

\bibitem[{Azadkia and Chatterjee(2021)}]{azadkia:2021}
Azadkia, M.; and Chatterjee, S. 2021.
\newblock {A simple measure of conditional dependence}.
\newblock \emph{The Annals of Statistics}, 49(6): 3070 -- 3102.

\bibitem[{Bartlett, Desmond, and Ferreira(2023)}]{esr_bartlett23}
Bartlett, D.~J.; Desmond, H.; and Ferreira, P.~G. 2023.
\newblock Exhaustive symbolic regression.
\newblock \emph{IEEE Transactions on Evolutionary Computation}.

\bibitem[{Burlacu, Kronberger, and Kommenda(2020)}]{operon_burlacu}
Burlacu, B.; Kronberger, G.; and Kommenda, M. 2020.
\newblock Operon C++: An Efficient Genetic Programming Framework for Symbolic Regression.
\newblock In \emph{Proceedings of the 2020 Genetic and Evolutionary Computation Conference Companion}, GECCO '20, 1562–1570.

\bibitem[{Chatterjee(2021)}]{chatterjee:2020}
Chatterjee, S. 2021.
\newblock A New Coefficient of Correlation.
\newblock \emph{Journal of the American Statistical Association}, 116(536): 2009--2022.

\bibitem[{Cranmer(2023)}]{cranmer2023pysr}
Cranmer, M. 2023.
\newblock Interpretable Machine Learning for Science with PySR and SymbolicRegression.jl.

\bibitem[{Deb, Ghosal, and Sen(2020)}]{deb:2020}
Deb, N.; Ghosal, P.; and Sen, B. 2020.
\newblock Measuring Association on Topological Spaces Using Kernels and Geometric Graphs.

\bibitem[{Feynman, Leighton, and Sands(2011)}]{feynman2011}
Feynman, R.~P.; Leighton, R.~B.; and Sands, M. 2011.
\newblock \emph{The Feynman Lectures on Physics, Vol. II: The New Millennium Edition: Mainly Electromagnetism and Matter}.
\newblock The Feynman Lectures on Physics. Basic Books.

\bibitem[{Guimerà et~al.(2020)Guimerà, Reichardt, Aguilar-Mogas, Massucci, Miranda, Pallarès, and Sales-Pardo}]{bayesianscientist_guimera20}
Guimerà, R.; Reichardt, I.; Aguilar-Mogas, A.; Massucci, F.~A.; Miranda, M.; Pallarès, J.; and Sales-Pardo, M. 2020.
\newblock A Bayesian machine scientist to aid in the solution of challenging scientific problems.
\newblock \emph{Science Advances}, 6(5): eaav6971.

\bibitem[{Holland(1975)}]{Holland:1975}
Holland, J.~H. 1975.
\newblock \emph{Adaptation in Natural and Artificial Systems}.
\newblock University of Michigan Press.
\newblock Second edition, 1992.

\bibitem[{Jin et~al.(2020)Jin, Fu, Kang, Guo, and Guo}]{jin2020bayesian}
Jin, Y.; Fu, W.; Kang, J.; Guo, J.; and Guo, J. 2020.
\newblock Bayesian Symbolic Regression.
\newblock arXiv:1910.08892.

\bibitem[{Kahlmeyer et~al.(2024)Kahlmeyer, Giesen, Habeck, and Voigt}]{kahlmeyer2024udfs}
Kahlmeyer, P.; Giesen, J.; Habeck, M.; and Voigt, H. 2024.
\newblock Scaling Up Unbiased Search-based Symbolic Regression.
\newblock In Larson, K., ed., \emph{Proceedings of the Thirty-Third International Joint Conference on Artificial Intelligence, {IJCAI-24}}, 4264--4272. International Joint Conferences on Artificial Intelligence Organization.
\newblock Main Track.

\bibitem[{Kamienny et~al.(2022)Kamienny, d'Ascoli, Lample, and Charton}]{transformer_kamienny22}
Kamienny, P.-A.; d'Ascoli, S.; Lample, G.; and Charton, F. 2022.
\newblock End-to-end Symbolic Regression with Transformers.
\newblock In Oh, A.~H.; Agarwal, A.; Belgrave, D.; and Cho, K., eds., \emph{Advances in Neural Information Processing Systems}.

\bibitem[{Kammerer et~al.(2020)Kammerer, Kronberger, Burlacu, Winkler, Kommenda, and Affenzeller}]{rationals_kammerer20}
Kammerer, L.; Kronberger, G.; Burlacu, B.; Winkler, S.~M.; Kommenda, M.; and Affenzeller, M. 2020.
\newblock Symbolic regression by exhaustive search: Reducing the search space using syntactical constraints and efficient semantic structure deduplication.
\newblock \emph{Genetic programming theory and practice XVII}, 79--99.

\bibitem[{Kommenda et~al.(2020)Kommenda, Burlacu, Kronberger, and Affenzeller}]{kommenda2020parameter}
Kommenda, M.; Burlacu, B.; Kronberger, G.; and Affenzeller, M. 2020.
\newblock Parameter identification for symbolic regression using nonlinear least squares.
\newblock \emph{Genetic Programming and Evolvable Machines}, 21(3): 471--501.

\bibitem[{Koza(1994)}]{koza1994genetic}
Koza, J.~R. 1994.
\newblock Genetic programming as a means for programming computers by natural selection.
\newblock \emph{Statistics and computing}, 4: 87--112.

\bibitem[{La~Cava et~al.(2021)La~Cava, Orzechowski, Burlacu, de~Franca, Virgolin, Jin, Kommenda, and Moore}]{srbench_lacava21}
La~Cava, W.; Orzechowski, P.; Burlacu, B.; de~Franca, F.; Virgolin, M.; Jin, Y.; Kommenda, M.; and Moore, J. 2021.
\newblock Contemporary Symbolic Regression Methods and their Relative Performance.
\newblock In Vanschoren, J.; and Yeung, S., eds., \emph{Proceedings of the Neural Information Processing Systems Track on Datasets and Benchmarks}, volume~1. Curran.

\bibitem[{La~Cava, Spector, and Danai(2016)}]{eplex_lacava16}
La~Cava, W.; Spector, L.; and Danai, K. 2016.
\newblock Epsilon-Lexicase Selection for Regression.
\newblock In \emph{Proceedings of the Genetic and Evolutionary Computation Conference 2016}, GECCO '16, 741–748. New York, NY, USA: Association for Computing Machinery.
\newblock ISBN 9781450342063.

\bibitem[{Meurer et~al.(2017)Meurer, Smith, Paprocki, \v{C}ert\'{i}k, Kirpichev, Rocklin, Kumar, Ivanov, Moore, Singh, Rathnayake, Vig, Granger, Muller, Bonazzi, Gupta, Vats, Johansson, Pedregosa, Curry, Terrel, Rou\v{c}ka, Saboo, Fernando, Kulal, Cimrman, and Scopatz}]{sympy_meurer17}
Meurer, A.; Smith, C.~P.; Paprocki, M.; \v{C}ert\'{i}k, O.; Kirpichev, S.~B.; Rocklin, M.; Kumar, A.; Ivanov, S.; Moore, J.~K.; Singh, S.; Rathnayake, T.; Vig, S.; Granger, B.~E.; Muller, R.~P.; Bonazzi, F.; Gupta, H.; Vats, S.; Johansson, F.; Pedregosa, F.; Curry, M.~J.; Terrel, A.~R.; Rou\v{c}ka, v.; Saboo, A.; Fernando, I.; Kulal, S.; Cimrman, R.; and Scopatz, A. 2017.
\newblock SymPy: symbolic computing in Python.
\newblock \emph{PeerJ Computer Science}, 3: e103.

\bibitem[{Mundhenk et~al.(2021)Mundhenk, Landajuela, Glatt, Santiago, Faissol, and Petersen}]{dsr2_petersen21}
Mundhenk, T.; Landajuela, M.; Glatt, R.; Santiago, C.; Faissol, D.; and Petersen, B. 2021.
\newblock Symbolic Regression via Neural-Guided Genetic Programming Population Seeding.

\bibitem[{Pearson(1920)}]{pearson:1920}
Pearson, K. 1920.
\newblock {Notes on the history of correlation}.
\newblock \emph{Biometrika}, 13(1): 25--45.

\bibitem[{Petersen et~al.(2021)Petersen, Larma, Mundhenk, Santiago, Kim, and Kim}]{dsr_petersen21}
Petersen, B.~K.; Larma, M.~L.; Mundhenk, T.~N.; Santiago, C.~P.; Kim, S.~K.; and Kim, J.~T. 2021.
\newblock Deep symbolic regression: Recovering mathematical expressions from data via risk-seeking policy gradients.
\newblock In \emph{International Conference on Learning Representations}.

\bibitem[{Schmidt and Lipson(2009)}]{eureqa_lipson09}
Schmidt, M.; and Lipson, H. 2009.
\newblock Distilling Free-Form Natural Laws from Experimental Data.
\newblock \emph{Science}, 324(5923): 81--85.

\bibitem[{Spearman(1904)}]{spearman:1904}
Spearman, C. 1904.
\newblock The proof and measurement of association between two things.
\newblock \emph{The American journal of psychology}, 15(1): 72--101.

\bibitem[{Stephens(2016)}]{gplearn_stephens16}
Stephens, T. 2016.
\newblock Genetic Programming in Python, with a scikit-learn inspired API: gplearn.

\bibitem[{Udrescu et~al.(2020)Udrescu, Tan, Feng, Neto, Wu, and Tegmark}]{feynmanAI2_udrescu20}
Udrescu, S.-M.; Tan, A.; Feng, J.; Neto, O.; Wu, T.; and Tegmark, M. 2020.
\newblock AI Feynman 2.0: Pareto-optimal symbolic regression exploiting graph modularity.
\newblock In Larochelle, H.; Ranzato, M.; Hadsell, R.; Balcan, M.; and Lin, H., eds., \emph{Advances in Neural Information Processing Systems}, volume~33, 4860--4871. Curran Associates, Inc.

\bibitem[{Udrescu and Tegmark(2020)}]{feynmanAI_udrescu20}
Udrescu, S.-M.; and Tegmark, M. 2020.
\newblock AI Feynman: A physics-inspired method for symbolic regression.
\newblock \emph{Science Advances}, 6(16): eaay2631.

\bibitem[{Virgolin et~al.(2021)Virgolin, Alderliesten, Witteveen, and Bosman}]{virgolin2021improving}
Virgolin, M.; Alderliesten, T.; Witteveen, C.; and Bosman, P.~A. 2021.
\newblock Improving model-based genetic programming for symbolic regression of small expressions.
\newblock \emph{Evolutionary computation}, 29(2): 211--237.

\end{thebibliography}

\clearpage
\onecolumn
\appendix
\noindent
\rule{\textwidth}{0.3pt}
\begin{center}
  \textbf{\LARGE Dimension Reduction for Symbolic Regression (Supplementary Material)}
\end{center}
\rule{\textwidth}{0.3pt}

\section{Beam Search}
The core process of our work is a beam search. This beam search takes a given regression problem and uses functional dependence measures to create regression problems with fewer independent variables. 
A given symbolic regressor then tries to solve each of these problems, creating expressions in the coordinates of these new problems. For each expression, we then map those coordinates back to the coordinates of the original problem and take the expression with the best model fit.

In the following we will illustrate these two steps using the introductory example of Washburns Formula, where we replaced the variables with the generic $x_i$:
\[
y = \sqrt{\frac{x_1x_2x_3 \cos (x_4)}{2x_5}}\,.
\]

\paragraph{Transformed regression problems.}
Starting from the original regression problem at the root, the search tree consists of all possible substitutions from a limited budget that could be performed. At each node we have a transformed dataset with independent variables $\hat{x}_i$ and dependent variable $\hat{y}$.
Using functional dependence measures, we select the best nodes for the beam search. At last, we return the datasets at the nodes on the path that leads to the dataset with the highest functional dependence that we encountered during our search. Together with the substitutions that were performed on this path, we also track how the variables translate to the variables of the original problem. 

\begin{table}[ht]
    \centering
    \begin{tabular}{ccccc}
         Step&Substitution&Type&Independent variables $\hat{x}$&Dependent variable $\hat{y}$\\
         \toprule
         1&&&$x_1,x_2,x_3,x_4,x_5$&$y$\\
         2&$g(\hat{x}_1, \hat{x}_2, \hat{x}_3) = \hat{x}_1\hat{x}_2\hat{x}_3$&input&$(x_1x_2x_3), x_4, x_5$&$y$\\
         3&$h(\hat{x_1}, \hat{y}) = \hat{y}/\sqrt{\hat{x_1}}$&out-input&$x_4, x_5$&$y/\sqrt{(x_1x_2x_3)}$\\
         4&$h(\hat{x_2}, \hat{y}) = \hat{y}\sqrt{\hat{x_2}}$&out-input&$x_4$&$y\sqrt{x_5/(x_1x_2x_3)}$\\
    \end{tabular}
    \caption{Example for the result of a beam search on Washburns formula. At each step, we create a new regression problem where the goal is to find a function $\hat{f}$ with $\hat{y} = \hat{f}(\hat{x})$.}
    \label{tab:example_beam_result}
\end{table}

In our example, such a path of substitutions and their translations is illustrated in Table~\ref{tab:example_beam_result}.
We have created four regression problems, where we know how the variables translate into the variables of the original regression problem.

\paragraph{Solving regression problems.}
The regression problems are now passed to a symbolic regressor, that tries to solve them. Each found expression is translated back into the coordinates of the original problem and solved for $y$.

Lets imagine a symbolic regressor takes problem number four from Table~\ref{tab:example_beam_result} and returns the function
\begin{align*}
    \hat{y} = \sqrt{\cos(\hat{x}_1)/2}\,.
\end{align*}
Then we can translate each variable back into the variables of the original problem:
\begin{align*}
    y\sqrt{x_5/(x_1x_2x_3)} = \sqrt{\cos(x_4)/2}\,.
\end{align*}
Now we solve this equation for $y$ using the computer algebra system SymPy.
\begin{align*}
    y = \sqrt{x_1x_2x_3 \cos (x_4)/(2x_5)}\,.
\end{align*}
 Note how this corresponds to finding $f_g$ for the performed out-input substitutions. 
The symbolic regressor iteratively generates expressions for each of the regression problems and translates them into the original coordinates. Finally the translated expression with the best model fit is returned (see section~\ref{subsec:performance}).

\section{Regression Datasets}
In this work we use two sets of regression problems with known ground truth formulas. Both datasets can be found in \href{https://github.com/kahlmeyer94/SREquations}{\textcolor{blue}{this repository}}.

\paragraph{Feynman.}
The Feynman problems consist of 95 formulas with samples from the famous Feynman Lectures \cite{feynman2011} and were introduced by \cite{feynmanAI_udrescu20} to test their symbolic regressor. It has since been used and established by \cite{srbench_lacava21} in their comprehensive benchmark suite.
The database can be found \href{https://space.mit.edu/home/tegmark/aifeynman.html}{\textcolor{blue}{here}}. It consists of expressions together with samples. 

\paragraph{Wikipedia.}
In their paper, \citet{bayesianscientist_guimera20} use 4077 formulas scraped from Wikipedias \href{https://en.wikipedia.org/wiki/List_of_scientific_equations_named_after_people}{\textcolor{blue}{list of eponymous equations}} to derive a prior distribution on natural formulas. In a first step, we compiled those formulas down to 897 regression problems by bringing them into a unified notation using variables $x_i$ and removing duplicates. For each of the expressions, we then selected 1000 sample points at which the corresponding function is valid using the following procedure:
Start with $c = 1$ and no points. As long as there are less than 1000 points, try to sample the remaining number of points from the interval $[-c, c]$. Only accept points, at which the function evaluates. If there are still points missing, increase the interval borders $c\leftarrow c+0.5$. If $c>150$ and there are still points missing, reject this expression.
With this procedure we only had to reject 17 problems and end up with a collection of 880 regression problems.

\paragraph{Noise.}
We evaluated the performance of different functional dependence measures on regression datasets with increasing levels of noise.
More specifically, for an independent variable $y$, we used additive gaussian noise with noise level $\gamma$ similar to \citet{srbench_lacava21}:
\begin{align*}
    y_\gamma = y + \varepsilon_\gamma\text{, where } \varepsilon_\gamma\sim\mathcal{N}\left(0, \gamma\sqrt{\frac{1}{n}\sum_{i=1}^n y_i^2}\right)\,.
\end{align*}

\section{Reduction Rate}
In our experiments, we used reduction rate to quantify how far the beam search is able to reduce the number of independent variables.
In order to calculate the reduction rate, we need to verify if a given substitution is indeed a substitution. 
If we know the true function $y = f(x)$, then we verify the correctness using the computer algebra system SymPy~\citep{sympy_meurer17}. For better illustration, we explain the verification process using the example function 
\begin{align*}
    y = f(x) = x_1x_2 + x_3\,.
\end{align*}

\paragraph{Input substitutions.} 
A valid input substitution for the example problem would be $g(x_1, x_2) = x_1x_2$.
To verify that $g$ is indeed a valid input substitution, we first introduce a new variable $\gamma = g(x_I)$ and solve for one of the $x_i\in x_I$.
In our example this can be done with
\begin{align*}
    x_1 = \frac{\gamma}{x_2}\,.
\end{align*}
Next, this expression is substituted into the known, original function $f(x)$:
\begin{align*}
    f(x) = x_1x_2 + x_3 = \frac{\gamma}{x_2}x_2 + x_3 = \gamma + x_3\,.
\end{align*}
We then use SymPy to verify that the resulting expression $\gamma + x_3$ does not depend on $x_I = \{x_1,x_2\}$ anymore.

\paragraph{Out-input substitutions.}
A valid out-input substitution for the example problem would be $h(x_3, y) = y - x_3$.
To verify, that $h$ is indeed a valid out-input substitution, we substitute the known expression $y = f(x)$ into $h$:
\begin{align*}
    h(x_3, y) = y - x_3 = (x_1x_2 + x_3) - x_3 = x_1x_2\,.
\end{align*}
We then use SymPy to verify that the resulting expression $x_1x_2$ does not depend on $x_I = \{x_3\}$ anymore.

\section{Symbolic Regression}
Symbolic regressors struggle with complex expressions, hence we introduced substitutions to reduce the number of independent variables and thus the complexity of the target expression.
In order to evaluate if symbolic regressors benefit from this reduction, we tested several regressors on different performance measures with and without the beam search.
Again, for all symbolic checks we used the SymPy~\citep{sympy_meurer17} package.

\subsection{Regressors}
We demonstrate the effect of using substitutions on five symbolic regressors that represent fundamentally different approaches to symbolic regression: systematic search, samplers and genetic algorithms.

\paragraph{Genetic Programming}
Most symbolic regression algorithms adhere to the \emph{genetic programming} paradigm~\citep{koza1994genetic,Holland:1975}. 
The core concept involves generating a population of expression trees, estimating their parameters, and to recombine the best-performing ones through subtree exchanges in order to create a new population. This approach has been implemented in the pioneering Eureqa system by~\citet{eureqa_lipson09} and in gplearn by \citet{gplearn_stephens16}. More recent implementations include~\citet{eplex_lacava16,kommenda2020parameter,virgolin2021improving}. 
From this class of regressors, we selected \textbf{Operon}~\citep{operon_burlacu} as a classical representative of this class.
Additionally, we chose \textbf{PySR}~\citep{cranmer2023pysr} as another representative that extends the classical evolutionary approach with an outer loop to multiple populations which occasionally exchange individuals.

\paragraph{Samplers.}
A different approach is to optimize the parameters of a \emph{sampler} over the space of symbolic expressions.
In the Bayesian inference approach, MCMC (Markov Chain Monte Carlo) is used for sampling expressions from a posterior distribution. \cite{jin2020bayesian} use a hand-designed prior distribution on expression trees, whereas \cite{bayesianscientist_guimera20} compile a prior distribution from a corpus of 4,077 mathematical expressions that have been extracted from Wikipedia articles on problems from physics and the social sciences. We select two approaches, that use deep learning to parameterize samplers for expression trees in preorder:
\textbf{DSR} by \citep{dsr_petersen21} and \cite{dsr2_petersen21} train a recurrent neural network (RNN) using reinforcement learning for each regression problem. Hyperparameters influencing the computational cost are the number of samples and the batchsize for training the RNN. \citet{transformer_kamienny22} use a \textbf{transformer} in an end-to-end fashion on a large data set of regression problems to translate regression tasks, given as a sequence of input output pairs, into a preorder sequence.
Contrary to the other symbolic regressors, the transformer is trained on a large set of regression problems before it is applied. Thus, at inference time, the transformer only needs to search over the learned distribution of tokens. This search is performed using beam search. Inside the beam search, the next tokens are either selected based on the highest probabilities or by sampling from the distribution. As pointed out by the authors, sampling is generally preferred as it produces more diverse candidate solutions. 

\paragraph{Systematic Search.}
Overall, the idea of an unbiased, more or less assumption-free \emph{systematic search} has shown the greatest potential when it comes to recovering expressions from data. Since the search space grows exponentially, these methods introduce techniques to move the target expressions into the search space that can be covered with a limited budget. 
The AIFeynman project by \citet{feynmanAI_udrescu20,feynmanAI2_udrescu20} uses a set of neural-network-based statistical property tests on the original regression problem. Statistically significant invariants are then used to identify valid substitutions that reduce the dimensionality of the regression problem. Other examples for systematic search are \citet{esr_bartlett23} and \citet{rationals_kammerer20}.
We selected \textbf{UDFS} by \citet{kahlmeyer2024udfs}, an unbiased search for expressions represented by directed, acyclic graphs. Their search is boosted by the use of variable augmentations, that is, additional coordinates in which the target regression problem becomes easier. These augmentations are identified and selected using a fixed hypothesis class (e.g. polynomials) under which the augmented problem is approximated. The computational cost for this DAG search is defined mainly by the maximum number of DAG skeletons and the number of intermediary nodes (no input or output nodes) inside the DAG.

\paragraph{Hyperparameters.}
\begin{table}[ht]
    \centering
    \begin{tabular}{lll}
    Regressor&Parameter&Value\\
    \toprule
    UDFS&Intermediary Nodes&3\\
    &Max skeletons&100.000\\
    \midrule
    DSR&Samples&50.000\\
    &Batchsize&1000\\
    \midrule
    Transformer&Beam Type&sampling\\
    &Beam Size&20\\
    \midrule
    PySR&&default\\
    \midrule
    Operon&&default
    \end{tabular}
    \caption{Hyperparameters chosen for symbolic regressors. Other parameters were left at default values.}
    \label{tab:hyperparams}
\end{table}
Our goal is not to compare the regressors against each other but to show the effect of adding substitutions. Hence we selected the hyperparameters to enable a fast search process of the regressors.
The selected hyperparameters are listed in Table~\ref{tab:hyperparams}.

\subsection{Performance Measures}
\label{subsec:performance}

Following \citet{srbench_lacava21}, we evaluate the performance of symbolic regressors along the three dimensions recovery, model fit and model complexity.

\paragraph{Recovery.} 
Symbolic regressors should return interpretable models. That is, the symbolic form of the models should carry information about the true, underlying process. 
Recovery is a proxy for this interpretability and is defined as the percentage of correctly recovered ground truth models from a collection of regression problems. 
Similar to~\citet{srbench_lacava21}, we consider a ground truth expression $f$ as recovered by a model $\hat{f}$, if either $f - \hat{f}$ can be symbolically resolved to a constant or $\hat{f}$ is non-zero and $\hat{f}/ f$ can be symbolically resolved to a constant.

\paragraph{Jaccard index.}
Recovery itself is an extremely strict measure, hence \citet{kahlmeyer2024udfs} suggested to use the Jaccard index as a soft version of recovery. The Jaccard index is a measure that compares the sets of subexpressions between a true expression and a model.
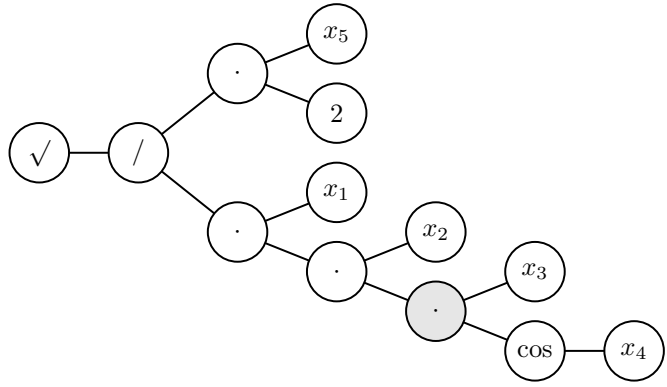
\begin{figure}[ht]
    \centering
    \tikzset{every picture/.style={line width=0.75pt}} %set default line width to 0.75pt        

\begin{tikzpicture}[x=0.75pt,y=0.75pt,yscale=-1,xscale=1]
%uncomment if require: \path (0,210); %set diagram left start at 0, and has height of 210

%Straight Lines [id:da1675292099823501] 
\draw    (25,85) -- (75,85) ;
%Straight Lines [id:da8249910847734852] 
\draw    (275,185) -- (325,185) ;
%Straight Lines [id:da05666620246552545] 
\draw    (225,165) -- (275,185) ;
%Straight Lines [id:da8254189356161471] 
\draw    (175,145) -- (225,165) ;
%Straight Lines [id:da7898427746997807] 
\draw    (175,145) -- (225,125) ;
%Straight Lines [id:da6097711716283495] 
\draw    (75,85) -- (125,125) ;
%Straight Lines [id:da4582451894621812] 
\draw    (75,85) -- (125,45) ;
%Straight Lines [id:da9322456779437011] 
\draw    (125,45) -- (175,65) ;
%Straight Lines [id:da9354061964946442] 
\draw    (125,45) -- (175,25) ;
%Shape: Circle [id:dp0006034762034256014] 
\draw  [fill={rgb, 255:red, 255; green, 255; blue, 255 }  ,fill opacity=1 ] (10,85) .. controls (10,76.72) and (16.72,70) .. (25,70) .. controls (33.28,70) and (40,76.72) .. (40,85) .. controls (40,93.28) and (33.28,100) .. (25,100) .. controls (16.72,100) and (10,93.28) .. (10,85) -- cycle ;
%Shape: Circle [id:dp27222230508456213] 
\draw  [fill={rgb, 255:red, 255; green, 255; blue, 255 }  ,fill opacity=1 ] (60,85) .. controls (60,76.72) and (66.72,70) .. (75,70) .. controls (83.28,70) and (90,76.72) .. (90,85) .. controls (90,93.28) and (83.28,100) .. (75,100) .. controls (66.72,100) and (60,93.28) .. (60,85) -- cycle ;
%Shape: Circle [id:dp9302617175171627] 
\draw  [fill={rgb, 255:red, 255; green, 255; blue, 255 }  ,fill opacity=1 ] (110,45) .. controls (110,36.72) and (116.72,30) .. (125,30) .. controls (133.28,30) and (140,36.72) .. (140,45) .. controls (140,53.28) and (133.28,60) .. (125,60) .. controls (116.72,60) and (110,53.28) .. (110,45) -- cycle ;
%Shape: Circle [id:dp923502221565487] 
\draw  [fill={rgb, 255:red, 255; green, 255; blue, 255 }  ,fill opacity=1 ] (160,25) .. controls (160,16.72) and (166.72,10) .. (175,10) .. controls (183.28,10) and (190,16.72) .. (190,25) .. controls (190,33.28) and (183.28,40) .. (175,40) .. controls (166.72,40) and (160,33.28) .. (160,25) -- cycle ;
%Shape: Circle [id:dp6228673701514125] 
\draw  [fill={rgb, 255:red, 255; green, 255; blue, 255 }  ,fill opacity=1 ] (160,65) .. controls (160,56.72) and (166.72,50) .. (175,50) .. controls (183.28,50) and (190,56.72) .. (190,65) .. controls (190,73.28) and (183.28,80) .. (175,80) .. controls (166.72,80) and (160,73.28) .. (160,65) -- cycle ;
%Straight Lines [id:da24177946932534544] 
\draw    (125,125) -- (175,145) ;
%Straight Lines [id:da15319301698658927] 
\draw    (125,125) -- (175,105) ;
%Shape: Circle [id:dp850839111129662] 
\draw  [fill={rgb, 255:red, 255; green, 255; blue, 255 }  ,fill opacity=1 ] (160,105) .. controls (160,96.72) and (166.72,90) .. (175,90) .. controls (183.28,90) and (190,96.72) .. (190,105) .. controls (190,113.28) and (183.28,120) .. (175,120) .. controls (166.72,120) and (160,113.28) .. (160,105) -- cycle ;
%Shape: Circle [id:dp7249796530893823] 
\draw  [fill={rgb, 255:red, 255; green, 255; blue, 255 }  ,fill opacity=1 ] (160,145) .. controls (160,136.72) and (166.72,130) .. (175,130) .. controls (183.28,130) and (190,136.72) .. (190,145) .. controls (190,153.28) and (183.28,160) .. (175,160) .. controls (166.72,160) and (160,153.28) .. (160,145) -- cycle ;
%Shape: Circle [id:dp38552437532549777] 
\draw  [fill={rgb, 255:red, 255; green, 255; blue, 255 }  ,fill opacity=1 ] (110,125) .. controls (110,116.72) and (116.72,110) .. (125,110) .. controls (133.28,110) and (140,116.72) .. (140,125) .. controls (140,133.28) and (133.28,140) .. (125,140) .. controls (116.72,140) and (110,133.28) .. (110,125) -- cycle ;
%Shape: Circle [id:dp9225369963581432] 
\draw  [fill={rgb, 255:red, 255; green, 255; blue, 255 }  ,fill opacity=1 ] (210,125) .. controls (210,116.72) and (216.72,110) .. (225,110) .. controls (233.28,110) and (240,116.72) .. (240,125) .. controls (240,133.28) and (233.28,140) .. (225,140) .. controls (216.72,140) and (210,133.28) .. (210,125) -- cycle ;
%Straight Lines [id:da8780948131044545] 
\draw    (225,165) -- (275,145) ;
%Shape: Circle [id:dp622435570469768] 
\draw  [fill={rgb, 255:red, 230; green, 230; blue, 230 }  ,fill opacity=1 ] (210,165) .. controls (210,156.72) and (216.72,150) .. (225,150) .. controls (233.28,150) and (240,156.72) .. (240,165) .. controls (240,173.28) and (233.28,180) .. (225,180) .. controls (216.72,180) and (210,173.28) .. (210,165) -- cycle ;
%Shape: Circle [id:dp6709208129158727] 
\draw  [fill={rgb, 255:red, 255; green, 255; blue, 255 }  ,fill opacity=1 ] (260,145) .. controls (260,136.72) and (266.72,130) .. (275,130) .. controls (283.28,130) and (290,136.72) .. (290,145) .. controls (290,153.28) and (283.28,160) .. (275,160) .. controls (266.72,160) and (260,153.28) .. (260,145) -- cycle ;
%Shape: Circle [id:dp8148224436839541] 
\draw  [fill={rgb, 255:red, 255; green, 255; blue, 255 }  ,fill opacity=1 ] (260,185) .. controls (260,176.72) and (266.72,170) .. (275,170) .. controls (283.28,170) and (290,176.72) .. (290,185) .. controls (290,193.28) and (283.28,200) .. (275,200) .. controls (266.72,200) and (260,193.28) .. (260,185) -- cycle ;
%Shape: Circle [id:dp037784979268315255] 
\draw  [fill={rgb, 255:red, 255; green, 255; blue, 255 }  ,fill opacity=1 ] (310,185) .. controls (310,176.72) and (316.72,170) .. (325,170) .. controls (333.28,170) and (340,176.72) .. (340,185) .. controls (340,193.28) and (333.28,200) .. (325,200) .. controls (316.72,200) and (310,193.28) .. (310,185) -- cycle ;

% Text Node
\draw (25,85) node    {$\sqrt{}$};
% Text Node
\draw (75,85) node    {$/$};
% Text Node
\draw (125,45) node    {$\cdot $};
% Text Node
\draw (175,25) node    {$x_{5}$};
% Text Node
\draw (175,65) node    {$2$};
% Text Node
\draw (175,105) node    {$x_{1}$};
% Text Node
\draw (175,145) node    {$\cdot $};
% Text Node
\draw (125,125) node    {$\cdot $};
% Text Node
\draw (225,125) node    {$x_{2}$};
% Text Node
\draw (225,165) node    {$\cdot $};
% Text Node
\draw (275,145) node    {$x_{3}$};
% Text Node
\draw (275,185) node    {$\cos$};
% Text Node
\draw (325,185) node    {$x_{4}$};

\end{tikzpicture}
    \caption{Expression tree corresponding to Washburns formula. If we take the gray node as root of a subtree, we get a subexpression $x_3\cos(x_4)$.}
    \label{fig:washburn_expression_tree}
\end{figure}

As subexpressions we consider all subtrees of a given expression tree. 
As an example, take the expression tree from the introductory example of Washburns formula, displayed in Figure~\ref{fig:washburn_expression_tree}. A subexpression corresponds to any subtree of the original expression tree.
In case of Washburns formula, if we take the lowest multiplication node (marked gray) as a root, the corresponding subexpression would be $x_3\cos(x_4)$.

In order to calculate the Jaccard index, we first used SymPy to create the sets of subexpressions $S$ and $\hat{S}$ for the simplified expression $f$ and the simplified model $\hat{f}$, respectively.
The Jaccard index then measures the similarity between those two sets
\[
J(S, \hat{S}) = \cfrac{|S\cap \hat{S}|}{|S\cup \hat{S}|}\,,
\]
as a number $\in[0, 1]$, with $J(S, \hat{S}) = 1$, if both expressions have exactly the same set of subexpressions.

\paragraph{Model fit.}
The standard measure for regression models is their ability to predict the values on held out test data
\[
\big(x^{(1)},y^{(1)}\big),\ldots,\big(x^{(n)},y^{(n)}\big)\,.
\]
During our experiments, we always used 20\% of the sample points uniformly selected as test data.

Let now $\hat{f}$ be a model. As a measure of model fit, we use the normalized root mean squared error (NRMSE):
\begin{align*}
    \text{NRMSE} = \sqrt{\cfrac{\sum_{i=1}^n\big(y^{(i)} - \hat{f}(x^{(i)})\big)^2}{\sum_{i=1}^n(y^{(i)})^2}}\,.
\end{align*}
The NRMSE ranges from $[0, \infty)$ and is zero, only if all predictions match the true values. Most importantly, the error is normalized. For different regression problems, the dependent variables $y$ lie in different ranges. With an unnormalized measure, regression problems with large $y$ would contribute more to the average model fit.

\iffalse
% Not used
Another closely related measure is the coefficient of determination, known as $R^2$-Score:
\begin{align*}
    R^2 = 1 - \cfrac{\sum_{i=1}^n\big(y^{(i)} - \hat{f}(x^{(i)})\big)^2}{\sum_{i=1}^n(y^{(i)}-\bar{y})^2}\,,
\end{align*}
where $\bar{y}$ denotes the mean value of all the $y^{(i)}$. The $R^2$-Score ranges from $(-\infty, 1]$ and is one only if the predictions match the true values. Depending on the variance of the true independent variables $y^{(i)}$, this model punishes prediction errors differently. Hence we instead used
\begin{align*}
    R^2_0 = \max(0, R^2)\,,
\end{align*}
as a more robust alternative.
\fi

\paragraph{Model complexity.}
To measure model complexity, we again follow \citet{srbench_lacava21} and count the number of nodes in the corresponding expression tree according to SymPy. 
The example expression in Figure~\ref{fig:washburn_expression_tree} has complexity 13.

\subsection{Results}
In Table~\ref{tab:performance_feynman} and Table~\ref{tab:performance_wiki}, we list the performance of the different symbolic regressors as well as a polynomial regression of degree two on the regression datasets.
For each regression dataset, we tracked the average performance over all regression problems of the base model (base) and the base model combined with the beam search (beam).
For both datasets, all base models significantly improve in recovery and Jaccard index. With the exception of the transformer, all models also improve in model fit, if substitutions are used. The model complexity does not change much but slightly increases on average. The only exception the polynomial regression, where the model complexity drops by more than 20 nodes on average.

Naturally, an increased recovery implies an increased Jaccard index and a better model fit. As a follow up, we also tracked the performances on those problems that could not be recovered. The results are displayed in Table~\ref{tab:performance_feynman_fail} and Table~\ref{tab:performance_wiki_fail}. 
It turns out, that even if problems could not be recovered, all algorithms have an increased Jaccard index, if they are combined with the beam search. That is, the expressions actually contain more parts of the true, underlying expression.

\begin{table}[ht]
    \centering
    \begin{tabular}{lllllllll}
    &\multicolumn{2}{c}{recovery}&\multicolumn{2}{c}{Jaccard}&\multicolumn{2}{c}{NRMSE}&\multicolumn{2}{c}{complexity}\\
    &base&beam&base&beam&base&beam&base&beam\\
    \toprule
    \textsc{UDFS}&0.58&\textbf{0.69}&0.61&\textbf{0.73}&0.08&\textbf{0.03}&\textbf{10.26}&12.71\\
    DSR&0.37&\textbf{0.66}&0.5&\textbf{0.67}&0.14&\textbf{0.03}&\textbf{19.09}&21.17\\
    \textsc{Transf}&0.01&\textbf{0.11}&0.15&\textbf{0.27}&0.09&\textbf{0.03}&33.53&\textbf{28.97}\\
    \textsc{PySR}&0.51&\textbf{0.62}&0.62&\textbf{0.71}&0.09&\textbf{0.03}&\textbf{11.74}&14.86\\
    \textsc{Operon}&0.41&\textbf{0.52}&0.38&\textbf{0.53}&0.05&\textbf{0.02}&\textbf{34.81}&40.93\\
    \textsc{Poly}&0.06&\textbf{0.34}&0.17&\textbf{0.46}&0.18&\textbf{0.08}&54.2&\textbf{32.09}\\
    \end{tabular}
    \caption{Average performance measures on Feynman problems}
    \label{tab:performance_feynman}
\end{table}
\begin{table}[ht]
    \centering
    \begin{tabular}{lllllllll}
    &\multicolumn{2}{c}{recovery}&\multicolumn{2}{c}{Jaccard}&\multicolumn{2}{c}{NRMSE}&\multicolumn{2}{c}{complexity}\\
    &base&beam&base&beam&base&beam&base&beam\\
    \toprule
    \textsc{UDFS}&0.56&\textbf{0.7}&0.61&\textbf{0.69}&0.49&\textbf{0.26}&\textbf{9.68}&11.78\\
    DSR&0.4&\textbf{0.61}&0.49&\textbf{0.61}&1.01&\textbf{0.54}&\textbf{14.6}&17.18\\
    \textsc{Transf}&0.09&\textbf{0.24}&0.2&\textbf{0.31}&\textbf{0.42}&0.82&35.21&\textbf{32.28}\\
    \textsc{PySR}&0.45&\textbf{0.54}&0.54&\textbf{0.59}&0.57&\textbf{0.27}&\textbf{11.5}&13.83\\
    \textsc{Operon}&0.35&\textbf{0.46}&0.34&\textbf{0.47}&1.52&\textbf{1.48}&\textbf{34.17}&36.92\\
    \textsc{Poly}&0.15&\textbf{0.35}&0.24&\textbf{0.44}&0.97&\textbf{0.69}&44.63&\textbf{30.1}\\
    \end{tabular}
    \caption{Average performance measures on Wikipedia problems.}
    \label{tab:performance_wiki}
\end{table}
\begin{table}[ht]
    \centering
    \begin{tabular}{lllllll}
    &\multicolumn{2}{c}{Jaccard}&\multicolumn{2}{c}{NRMSE}&\multicolumn{2}{c}{complexity}\\
    &base&beam&base&beam&base&beam\\
    \toprule
    \textsc{UDFS}&0.25&\textbf{0.27}&0.16&\textbf{0.09}&\textbf{13.41}&20.14\\
    DSR&0.2&\textbf{0.22}&0.22&\textbf{0.1}&\textbf{24.03}&35.91\\
    \textsc{Transf}&0.14&\textbf{0.22}&0.08&\textbf{0.04}&33.61&\textbf{30.99}\\
    \textsc{PySR}&0.26&\textbf{0.27}&0.18&\textbf{0.09}&\textbf{15.0}&22.72\\
    \textsc{Operon}&0.12&\textbf{0.14}&0.07&\textbf{0.05}&\textbf{56.77}&70.43\\
    \textsc{Poly}&0.11&\textbf{0.2}&0.19&\textbf{0.12}&59.9&\textbf{39.62}\\
    \end{tabular}
    \caption{Average performance measures on unrecovered Feynman problems.}
    \label{tab:performance_feynman_fail}
\end{table}
\begin{table}[ht]
    \centering
    \begin{tabular}{lllllll}
    &\multicolumn{2}{c}{Jaccard}&\multicolumn{2}{c}{NRMSE}&\multicolumn{2}{c}{complexity}\\
    &base&beam&base&beam&base&beam\\
    \toprule
    \textsc{UDFS}&0.23&\textbf{0.26}&0.93&\textbf{0.83}&\textbf{12.07}&17.52\\
    DSR&0.2&\textbf{0.24}&2.07&\textbf{1.35}&\textbf{18.55}&26.64\\
    \textsc{Transf}&0.14&\textbf{0.17}&\textbf{0.55}&1.11&39.59&\textbf{38.28}\\
    \textsc{PySR}&0.21&\textbf{0.22}&1.04&\textbf{0.57}&\textbf{13.93}&18.71\\
    \textsc{Operon}&0.13&\textbf{0.15}&\textbf{2.39}&2.77&\textbf{48.79}&60.31\\
    \textsc{Poly}&0.11&\textbf{0.17}&1.11&\textbf{1.05}&47.01&\textbf{36.66}\\
    \end{tabular}
    \caption{Average performance measures on unrecovered Wikipedia problems.}
    \label{tab:performance_wiki_fail}
\end{table}

\end{document}